\documentclass{article} 
\usepackage{nfam2026_workshop}
\usepackage{times}
\usepackage{graphicx}
\usepackage{amsmath,amssymb,amsfonts,bm}

\usepackage{amsmath,amsfonts,bm}

\def\eqref#1{equation~\ref{#1}}

\def\1{\bm{1}}

\DeclareMathAlphabet{\mathsfit}{\encodingdefault}{\sfdefault}{m}{sl}
\SetMathAlphabet{\mathsfit}{bold}{\encodingdefault}{\sfdefault}{bx}{n}

\usepackage{hyperref}
\usepackage{url}

\title{Thermal Robustness of Retrieval in Dense\\ Associative Memories: LSE vs LSR Kernels}

\author{Tatiana Petrova \\
Interdisciplinary Centre for Security, Reliability and Trust (SnT) \\
University of Luxembourg \\
\texttt{tatiana.petrova@uni.lu}
}

\iclrfinalcopy 
\begin{document}

\maketitle

\begin{abstract}
Understanding whether retrieval in dense associative memories survives thermal noise is essential for bridging zero-temperature capacity proofs with the finite-temperature conditions of practical inference and biological computation. We use Monte Carlo simulations to map the retrieval phase boundary of two continuous dense associative memories (DAMs) on the $N$-sphere with an exponential number of stored patterns $M = e^{\alpha N}$: a log-sum-exp (LSE) kernel and a log-sum-ReLU (LSR) kernel. Both kernels share the zero-temperature critical load $\alpha_c(0)=0.5$, but their finite-temperature behavior differs markedly. The LSE kernel sustains retrieval at arbitrarily high temperatures for sufficiently low load, whereas the LSR kernel exhibits a finite support threshold below which retrieval is perfect at any temperature; for typical sharpness values this threshold approaches $\alpha_c$, making retrieval nearly perfect across the entire load range. We also compare the measured equilibrium alignment with analytical Boltzmann predictions within the retrieval basin.
\end{abstract}

\section{Introduction}
\label{sec:intro}
Associative memory networks store patterns for content-addressable retrieval from noisy or partial cues. The classical Hopfield network \cite{Hopfield1982} reliably retrieves up to approximately $0.14N$ patterns for $N$ binary neurons \cite{amit1985storing}. Dense associative memories (DAMs) with higher-order or exponential interactions achieve dramatically larger capacities: \citet{demircigil2017model} proved $M_{\max} \propto e^{\alpha N}$ with $\alpha \approx 0.35$ for binary states, and \citet{ramsauer2021hopfield} extended this to continuous states on the $N$-sphere using a log-sum-exp (LSE) energy, proving $\alpha_c = 0.5$ at zero temperature. The gradient dynamics of this energy correspond to softmax attention in Transformers, making DAMs a theoretical framework for attention-like retrieval. More recently, \citet{hoover2025dense} proposed the log-sum-ReLU (LSR) energy based on the compactly supported Epanechnikov kernel, achieving the same exponential capacity without exponential interactions.

Understanding the finite-temperature behavior of these models is important for at least two reasons. First, temperature serves as a natural proxy for noise in biological and hardware implementations: any physical realization of associative retrieval operates at effectively nonzero temperature due to stochastic fluctuations in the dynamics. Second, the connection between DAM energy minimization and softmax attention means that the retrieval phase boundary directly characterizes the robustness of attention-like computation. Temperature provides a principled measure of the perturbation that attention-like retrieval can tolerate. Despite this practical relevance, existing capacity results are almost exclusively stated at zero temperature, leaving a significant gap between theory and the conditions under which these models would actually operate.

While zero-temperature capacity is now well established, it provides an incomplete picture: it guarantees that a retrieval fixed point exists but says nothing about whether it survives thermal fluctuations—precisely the regime relevant to noisy inference, sampling-based decoding in language models, and stochastic neural dynamics. At finite temperature $T$, retrieval is governed by a competition between energetic alignment with a target memory and the entropic cost of constraining the state on the sphere. Classical analyses of the Hopfield model showed that this interplay produces rich phase structure, including retrieval, spin-glass, and paramagnetic phases \cite{amit1987statistical}. Recent work has begun addressing thermodynamic aspects of modern DAMs \cite{lucibello2024exponential, rooke2026stochastic}, but the full retrieval phase boundary for continuous spherical DAMs has not been mapped numerically. In this work, we fill this gap by using Monte Carlo (MC) simulations to map, for the first time, the retrieval phase boundary in the temperature--load $(T, \alpha)$ plane for both the LSE and LSR kernels, and compare the measured equilibrium alignment with analytical Boltzmann predictions within a single retrieval basin.

We define both the network state $\bm{x}$ and the $M = e^{\alpha N}$ stored patterns $\bm{\xi}^\mu$ on the $N$-sphere, $\|\bm{x}\|^2 = \|\bm{\xi}^\mu\|^2 = N$. The energy is constructed via kernel aggregation over squared Euclidean distances to stored patterns:
\begin{equation}
    H_{\rm LSE}(\bm{x}) = -\frac{1}{\beta_{\rm net}} \ln \sum_{\mu=1}^M \exp\left[ - \frac{\beta_{\rm net}}{2} d^2(\bm{x}, \bm{\xi}^\mu) \right] \,,
    \label{eq:H_lse}
\end{equation}

\begin{equation}
    H_{\rm LSR}(\bm{x}) = -\frac{1}{\beta_{\rm net}} \ln  \sum_{\mu=1}^M \max\left[\epsilon,  1 - \frac{\beta_{\rm net}}{2} d^2(\bm{x}, \bm{\xi}^\mu) \right]  \,,
    \label{eq:H_lsr}
\end{equation}
where $\epsilon > 0$ prevents logarithmic singularities and $\beta_{\rm net}$ is the inverse squared kernel width, controlling the sharpness of the energy landscape. The squared Euclidean distance decomposes as
\begin{equation}
    d^2(\bm{x}, \bm{\xi}^\mu) = \|\bm{x}-\bm{\xi}^\mu\|^2 = 2N (1 - \phi^\mu) \,,
    \label{eq:edist}
\end{equation}
in terms of the \textit{alignment} of the state with pattern $\mu$:
\begin{equation}
    \phi^\mu(\bm{x}) = \frac{1}{N} \bm{x} \cdot \bm{\xi}^\mu\,,
    \label{eq:alignment}
\end{equation}

which equals $1$ at perfect retrieval ($\bm{x} = \bm{\xi}^\mu$) and concentrates near $0$ for uncorrelated states in high dimensions. The LSE kernel assigns nonzero weight to all $M$ patterns everywhere on the sphere (Gaussian, infinite support), while the LSR kernel truncates contributions from distant patterns (Epanechnikov, finite support). For the LSR kernel it is convenient to introduce the rescaled sharpness $b \equiv N\beta_{\rm net}$, which controls the support boundary: pattern $\mu$ contributes to the energy only when $\phi^\mu > 1 - 1/b$, requiring $b > 1$ for a nontrivial retrieval basin. A direct comparison between the two kernels at equal $\beta_{\rm net}$ corresponds to $b = N$.

The system is coupled to a thermal bath at temperature $T$, with equilibrium governed by the Boltzmann distribution $\propto \exp(-H/T)$. Our MC simulations reveal that: (i) both kernels share the zero-temperature critical load $\alpha_c(0) = 0.5$, but their finite-temperature behavior differs markedly; (ii) the LSE kernel sustains retrieval at arbitrarily high temperatures for sufficiently low load; and (iii) the LSR kernel exhibits a threshold $\alpha_{\rm th} = (1 - b^{-1})^2/2$ below which retrieval is perfect at any temperature, and a finite maximal retrieval temperature above this threshold.

\section{Monte Carlo Simulations}
\label{sec:mc}

We use Metropolis--Hastings (MH) Monte Carlo to test whether the retrieval basin is stable against thermal perturbations: generate random patterns, initialize the system near a target pattern, and evolve under the Boltzmann distribution at temperature~$T$. 
A key difficulty in simulating models~(\ref{eq:H_lse}) and~(\ref{eq:H_lsr}) is the exponentially large number of patterns $M \sim \exp(\alpha N)$. For $\alpha \in [0.01,\, 0.55]$, no single fixed~$N$ keeps $M$ both large enough at small~$\alpha$ and computationally tractable at large~$\alpha$. We therefore adopt an adaptive-$N$ scheme:
\begin{equation}
    N(\alpha) = \left\lfloor \frac{\ln M(\alpha)}{\alpha} \right\rceil,
\end{equation}
which keeps the number of patterns manageable across the entire range of~$\alpha$ at the cost of reducing~$N$.

We generate $N_{\rm tr}$ states in the vicinity of a pattern with random $\phi_{\mathrm{init}} \sim \mathrm{U}(0.75,\, 1.0)$:
\begin{equation}
    \mathbf{x}_{\mathrm{init}} = \phi_{\mathrm{init}}\, \boldsymbol{\xi}^1
    + \sqrt{1 - \phi_{\mathrm{init}}^2}\;\sqrt{N}\; \hat{\mathbf{u}}_\perp
    \label{eq:xinit}
\end{equation}
($\hat{\mathbf{u}}_\perp$ is a random unit vector orthogonal to $\boldsymbol{\xi}^1$), each serving as the initial condition for an independent trial. Each trial uses a freshly generated pattern set and is evolved for $n_{\rm eq} = 16{,}384$ equilibration steps (unmeasured), followed by $n_{\rm samp} = 4{,}096$ sampling steps over which the time-averaged alignment is computed. The final alignment is averaged over all $N_{\rm tr}$ trials.

The MH scheme consists of two steps: (i)~propose a new state
\begin{equation}
    \bm{x}' = \bm{x} + \sigma\,\bm{\eta}\,,\quad
    \bm{x}' \leftarrow \sqrt{N}\,\bm{x}'/\|\bm{x}'\|\,;\quad
    \bm{\eta} \sim \mathcal{N}(0, I_N)\,,\quad
    \sigma = 2.4\,T/\sqrt{N}\,,
\end{equation}
and (ii)~accept it with probability $\min(1,\, e^{-\Delta H / T})$, where $\Delta H = H(\bm{x}') - H(\bm{x})$ is the change in the Hamiltonian between the proposed and current states.
In the LSR case, proposals that fall outside the support of all patterns are automatically rejected, enforcing the hard-wall constraint, since $H = +\infty$.

Both kernels use the same simulation infrastructure. We scan $\alpha \in [0.01,\, 0.55]$ in steps of $0.01$ and $T \in [0.025,\, 2.0]$ in steps of $0.05$. The number of patterns $M(\alpha)$ ranges from $20{,}000$ to $500{,}000$ following a power law with exponent $\gamma = 10$, and all $M$ patterns contribute to the energy evaluation. The resulting dimension $N = \lfloor \ln M / \alpha \rceil$ ranges from ${\sim}\,990$ at $\alpha = 0.01$ to ${\sim}\,24$ at $\alpha = 0.55$. The number of independent trials $N_{\rm tr}$ decreases linearly from 512 to~64 as $\alpha$ increases. All temperatures and trials are batched on the GPU; the code groups $\alpha$ values into chunks fitting within a ${\sim}\,43$~GB memory budget, processing one chunk at a time.

\begin{figure*}[t]
  \vskip 0.1in
  \begin{center}
    \centerline{\includegraphics[width=0.99\textwidth]{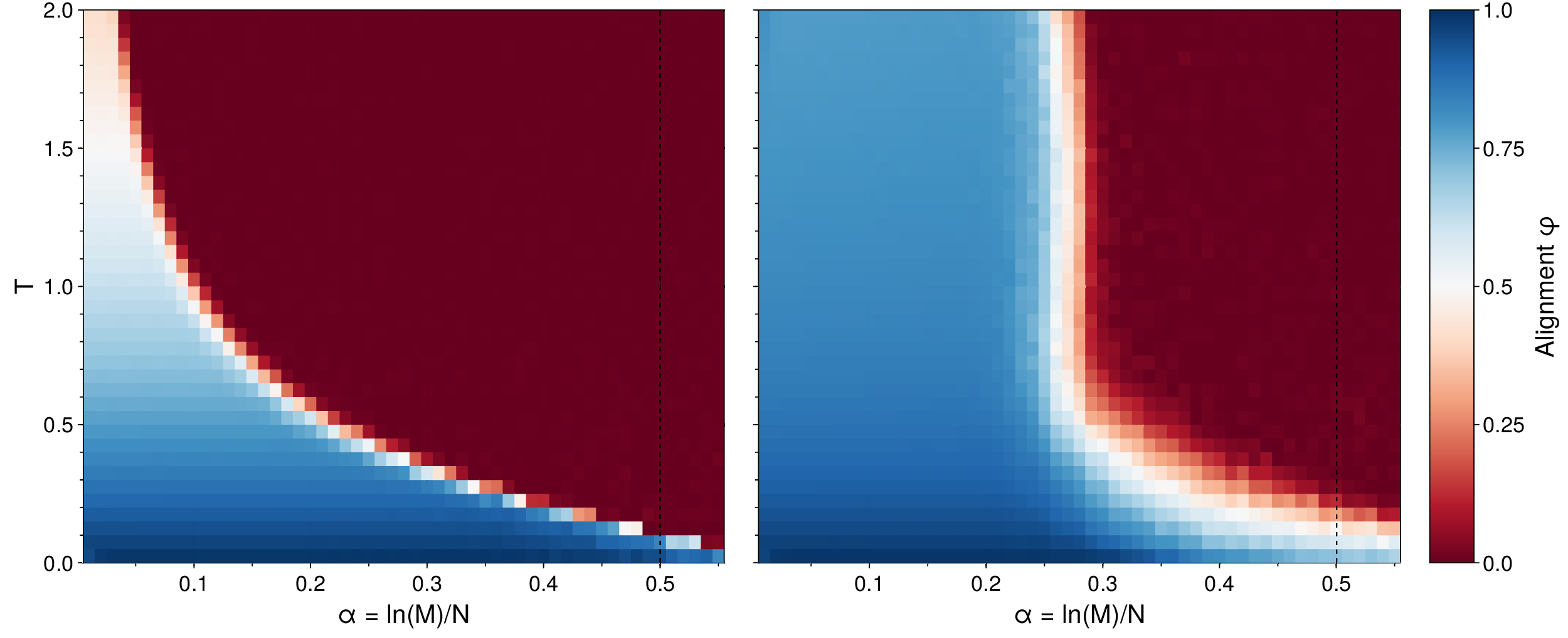}}
\caption{Phase diagrams for the spherical DAM with exponential capacity $M = e^{\alpha N}$: LSE (left, $\beta_{\mathrm{net}} = 1$) and LSR (right, $b=3.41$). The non-retrieval region is shown in red. The retrieval boundary lies below and to the left of the visible red boundary, corresponding to $\phi\sim 1$ at low~$T$ and decreasing to $\phi\sim 0.4$ at high~$T$ due to thermal broadening of the state (see main text). LSR exhibits a sharp threshold at $\alpha_{\text{th}} = (1-b^{-1})^2/2$, below which retrieval is perfect at any temperature. In both cases, retrieval breaks down at $\alpha = 0.5$ (dashed line).}
    \label{fig:phase_diagram}
  \end{center}
  \vskip -0.1in
\end{figure*}

The results of our Monte Carlo simulations are presented in Fig.\,\ref{fig:phase_diagram}. The left panel shows the $(\alpha,T)$-map of the average alignment for the LSE kernel with $\beta_{\mathrm{net}} = 1$. The red area corresponds to the non-retrieval phase, where the system explores the energy landscape extensively and the average alignment drops to zero. In the low-temperature limit, we reproduce the result of~\cite{ramsauer2021hopfield} that $M_{\rm max} \propto \exp(\alpha N)$ with $\alpha = 0.5$. At the low-$\alpha$ end, the pattern density is too low to create a dense net of local minima, so the average alignment is governed by the proximity to the retrieval pattern. However, its value can be far from unity due to thermal broadening.

The right panel shows the same map for the LSR kernel with $b = N\beta_{\mathrm{net}} = 3.41$. Because the kernel has finite support, spurious patterns with $\phi^\mu < \phi_{\rm c} = 1-1/b$ do not contribute to the energy landscape. For random patterns on the $N$-sphere, the overlap $\phi^\mu$ is approximately Gaussian with variance $1/N$, so the expected number of patterns exceeding $\phi_c$ becomes appreciable only when $\alpha > \alpha_{\text{th}} = (1-b^{-1})^2/2$. Accordingly, for $\alpha < \alpha_{\text{th}}$ the average alignment is determined solely by thermal broadening around the retrieval minimum and is independent of~$\alpha$, consistent with the panel. For $\alpha > \alpha_{\text{th}}$, spurious minima are close enough to the retrieval basin to allow thermally activated jumps between them, increasingly so at higher temperatures. One therefore expects the retrieval boundary to be a descending curve, starting below $T \simeq 0.5$ at $\alpha = 0.25$ and reaching zero at $\alpha = 0.5$. For large~$b$, e.g.\ corresponding to a fixed $\beta_{\mathrm{net}}$ as in the LSE case, one would expect the non-retrieval region to shrink to a strip  $\alpha \ge 0.5$.

A measured alignment $\phi < 1$ does not necessarily indicate retrieval failure. Even when the retrieval basin is the unique thermodynamic state (no local minima competition), thermal fluctuations reduce the alignment at finite temperature to a Boltzmann-averaged value $\phi_{\mathrm{eq}}(T) < 1$. We now evaluate this quantity. For the LSE kernel, the internal energy density of the single basin is $u(\phi) = 1 - \phi$, and the equilibrium alignment is given by
\begin{equation}\label{eq:boltzmann-avg}
    \phi_{\mathrm{eq}}(T) = \int_{-1}^{1} \phi\; n(\phi)\; e^{-N\,u(\phi)/T}\, \mathrm{d}\phi \;\bigg/\; \int_{-1}^{1} n(\phi)\; e^{-N\,u(\phi)/T}\, \mathrm{d}\phi\,,
\end{equation}
where $n(\phi)$ is the density of states at alignment~$\phi$, proportional to the surface area of an $(N\!-\!1)$-sphere:
\begin{equation}
    n(\phi) \propto [N (1 - \phi^2)]^{(N-2)/2}.
\end{equation}
For the LSR kernel, the same expression applies with $u(\phi) = -b^{-1}\ln\bigl[1 - b(1-\phi)\bigr]$ and the integration domain restricted to $\phi \in (\phi_c,\, 1]$.

\begin{figure}[t]
\centering
\includegraphics[width=0.45\columnwidth]{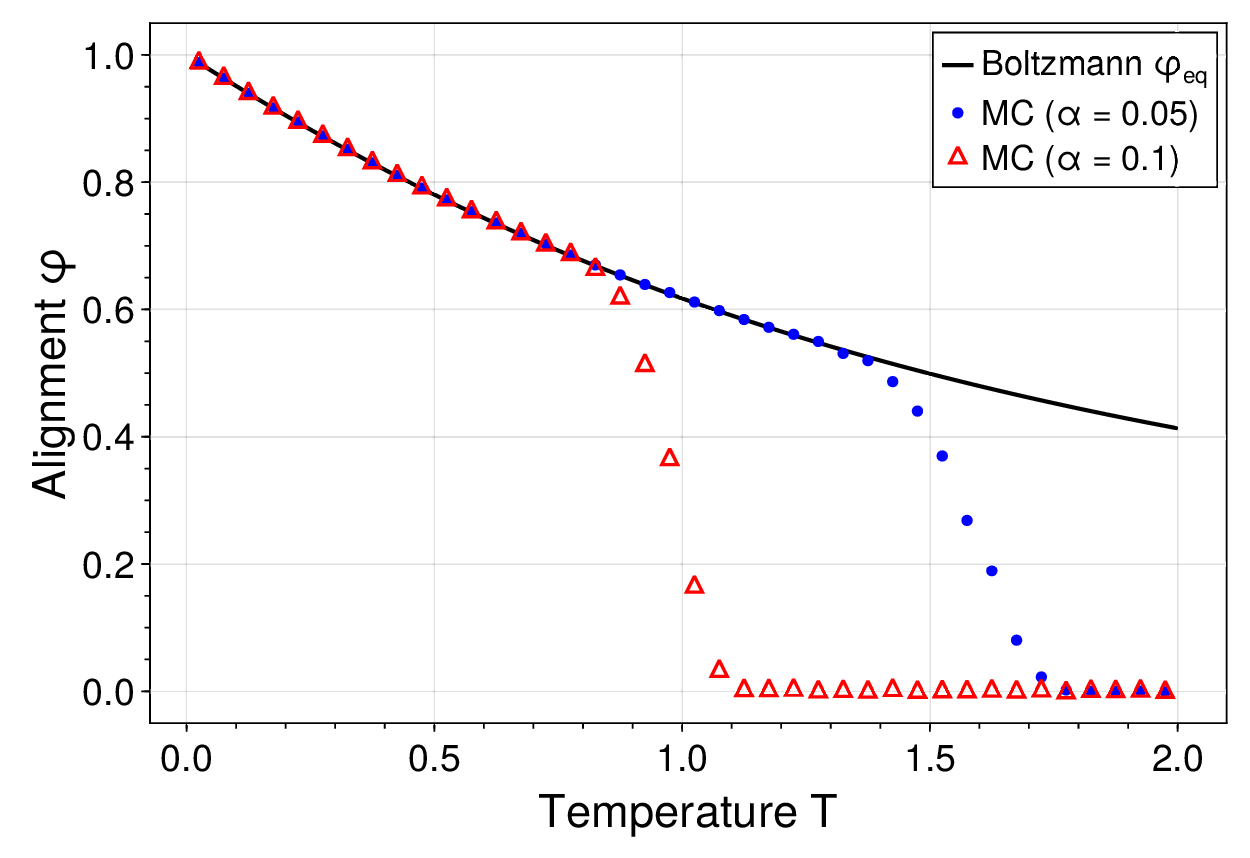} \hspace{5mm} \includegraphics[width=0.45\columnwidth]{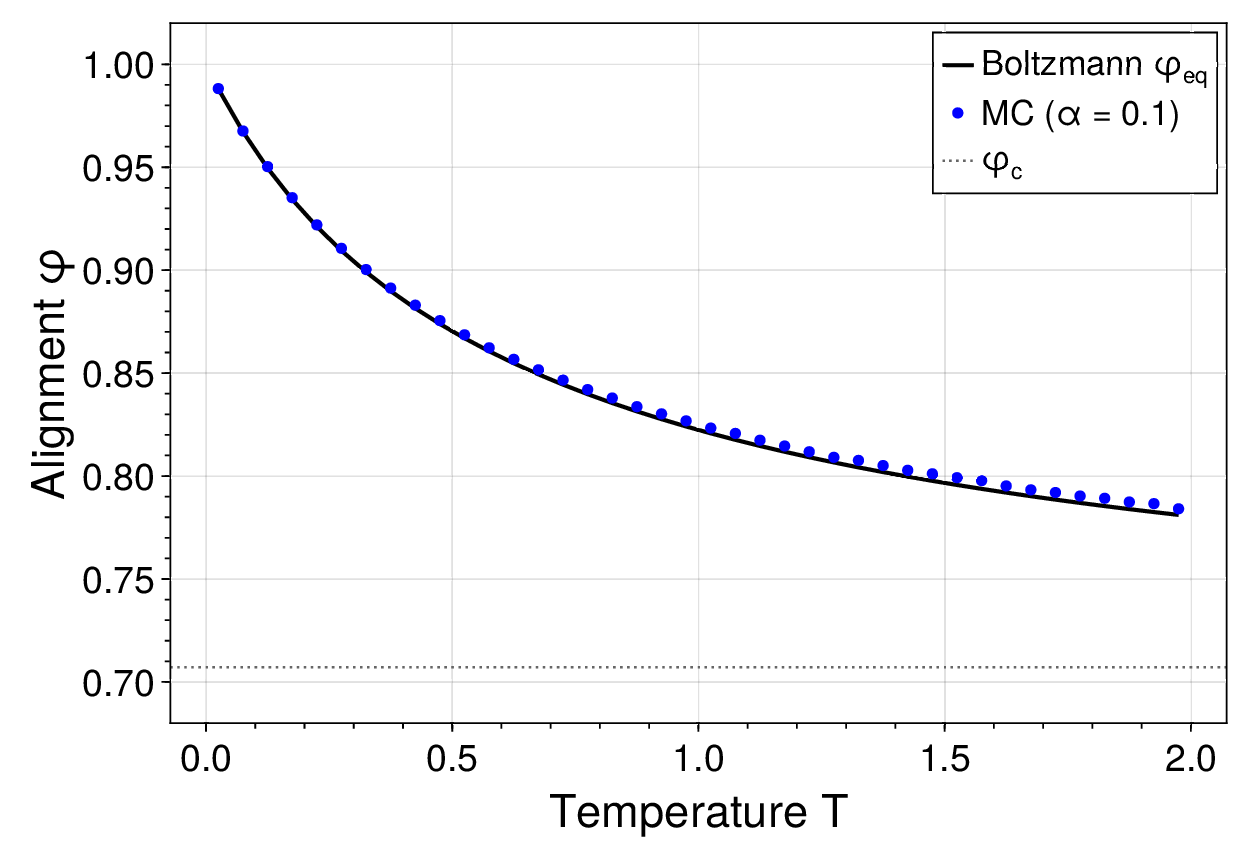}
\caption{Comparison of the Boltzmann equilibrium $\phi_{\rm eq}$ (solid curve) within the LSE basin with the Monte Carlo simulations at $\alpha = 0.05, 0.1$ (left), and the Boltzmann equilibrium $\phi_{\rm eq}$ within the LSR retrieval basin with the Monte Carlo simulations at $\alpha = 0.1$ (right). The
dotted line shows the hard-wall threshold $\phi_c$.}
\label{fig:boltzmann}
\end{figure}

Fig.\,\ref{fig:boltzmann} compares the Boltzmann-averaged $\phi_{\mathrm{eq}}(T)$ with Monte Carlo simulation results. The left panel shows the LSE kernel, with the average MC alignment at $\alpha = 0.05$ and $0.1$ (red triangles and blue dots, respectively). Before the retrieval transition, the data follow the equilibrium prediction, indicating that temperature alone can significantly reduce the alignment. Beyond the transition, the alignment drops to $\phi \approx 0$. The monotonic decrease of $\phi_{\mathrm{eq}}$ with~$T$ reflects the entropy--energy competition, even though the retrieval basin remains perfectly stable at these loads.
Similar behavior is observed for the LSR kernel (right panel) at $\alpha = 0.1$, which lies entirely within the retrieval phase.

\section{Summary, Discussion, and Outlook}
\label{sec:conclusions}
We have mapped the retrieval phase boundary of two continuous spherical DAMs with exponential capacity, $M = e^{\alpha N}$, using Monte Carlo simulations. Both the LSE and LSR kernels share the same zero-temperature critical load $\alpha_c = 0.5$, but their finite-temperature behavior differs qualitatively. The LSE kernel, with infinite support, sustains retrieval at arbitrarily high temperatures for sufficiently low $\alpha$. The LSR kernel, with finite support controlled by the rescaled sharpness $b = N\beta_{\mathrm{net}}$, exhibits a sharp threshold $\alpha_{\mathrm{th}} = (1 - b^{-1})^2/2$ below which retrieval is perfect at any temperature, and a maximal retrieval temperature above $\alpha_{\mathrm{th}}$. When compared at equal $\beta_{\mathrm{net}}$, i.e.\ $b = N \gg 1$, the LSR kernel shows substantially greater thermal robustness than LSE, since its finite support suppresses contributions from distant spurious patterns entirely.
Since the gradient dynamics of the LSE energy correspond to softmax attention, these results suggest that attention-based retrieval is inherently robust to stochastic perturbations at sub-critical load, a property that may help explain the empirical noise tolerance of Transformer architectures.

In the non-retrieval region (red area in Fig.\,\ref{fig:phase_diagram}), the average alignment drops to $\phi \approx 0$. Identifying the nature of this phase requires care. In the classical Hopfield model, \citet{amit1987statistical} identified three distinct non-retrieval states. The \emph{paramagnetic} phase occurs at high temperatures ($T > T_g$), where the system is fully disordered and ergodic. Below $T_g$, a \emph{spin-glass} phase appears, where the overlap with all stored patterns also vanishes but the system freezes into one of exponentially many random configurations unrelated to any stored pattern; this phase is distinguished from the paramagnetic one by a nonzero self-overlap $q > 0$~\citep{ea75}. A third class are the \emph{mixture states}, metastable configurations at low $\alpha$ and low temperature in which the system has finite overlap with several patterns simultaneously.

In our continuous spherical model, the present simulations do not distinguish between these scenarios. Measuring $q$ across the $(\alpha, T)$ plane and comparing with analytical mean-field predictions for these continuous models are natural next steps that would clarify the nature of the transitions and provide an independent check on the phase boundary location. On the computational side, the finite support of the LSR kernel suggests a possible improvement: near the target pattern, only $K \ll M$ patterns have overlap $\phi^\mu > \phi_c$ and contribute to the energy, with $K \sim \mathrm{Poisson}(M\, p_{\rm tail})$. Pre-selecting these active patterns would replace the $N \times M$ pattern array with an $N \times K$ array, raising the accessible dimension from $N \approx 24$ to $N \approx 50$ at the largest loads, though whether this approximation remains accurate as the state drifts from the target during MC evolution requires further investigation.

\subsubsection*{Acknowledgments}

\bibliography{dense_am}
\bibliographystyle{nfam2026_workshop}

\end{document}